\documentclass[10pt, a4paper]{article}
\usepackage{lrec2022} 
\usepackage{multibib}
\newcites{languageresource}{Language Resources}
\usepackage{microtype}
\usepackage{graphicx}
\usepackage{tabularx}
\usepackage{paralist}
\usepackage{soul}
\usepackage{balance}
\usepackage{titlesec}
\titleformat{\section}{\normalfont\large\bfseries\center}{\thesection.}{1em}{}
\titleformat{\subsection}{\normalfont\SmallTitleFont\bfseries\raggedright}{\thesubsection.}{1em}{}
\titleformat{\subsubsection}{\normalfont\normalsize\bfseries\raggedright}{\thesubsubsection.}{1em}{}
\renewcommand\thesection{\arabic{section}}
\renewcommand\thesubsection{\thesection.\arabic{subsection}}
\renewcommand\thesubsubsection{\thesubsection.\arabic{subsubsection}}

\usepackage{epstopdf}
\usepackage[utf8]{inputenc}

\usepackage[hidelinks]{hyperref}
\usepackage{xstring}
\usepackage{subcaption}
\usepackage{mwe}
\usepackage{color}

\newcommand{\ds}[1]{\textbf{\textcolor{red}{[Debjoy: #1]}}}

\title{Merkel Podcast Corpus: A Multimodal Dataset \\Compiled from 16 Years of Angela Merkel's Weekly Video Podcasts}

\name{Debjoy Saha$^{1,2}$, Shravan Nayak$^{1,3}$, Timo Baumann$^{1,4}$}

\address{$^1$Universität Hamburg, 
$^2$IIT Kharagpur,
$^3$IIT (BHU) Varanasi,
$^4$OTH Regensburg\\
sahadebjoy10@iitkgp.ac.in, 
pshravan.nayak.ece17@itbhu.ac.in, 
timo.baumann@oth-regensburg.de 
}

\abstract{
We introduce the Merkel Podcast Corpus, an audio-visual-text corpus in German collected from 16 years of (almost) weekly Internet podcasts of former German chancellor Angela Merkel. To the best of our knowledge, this is the first single speaker corpus in the German language consisting of audio, visual and text modalities of comparable size and temporal extent. 
We describe the methods used with which we have collected and edited the data which involves downloading the videos, transcripts and other metadata, forced alignment, performing active speaker recognition and face detection to finally curate the single speaker dataset consisting of utterances spoken by Angela Merkel. The proposed pipeline is general and can be used to curate other datasets of similar nature, such as talk show contents. Through various statistical analyses and applications of the dataset in talking face generation and TTS, we show the utility of the dataset. We argue that it is a valuable contribution to the research community, in particular, due to its realistic and challenging material at the boundary between prepared and spontaneous speech. 
 \\[5pt] \Keywords{multi-modal, German, single-speaker, corpus, speaker diarization, forced alignment, cross-modal learning} \\[5pt] 
    \includegraphics[width=\linewidth]{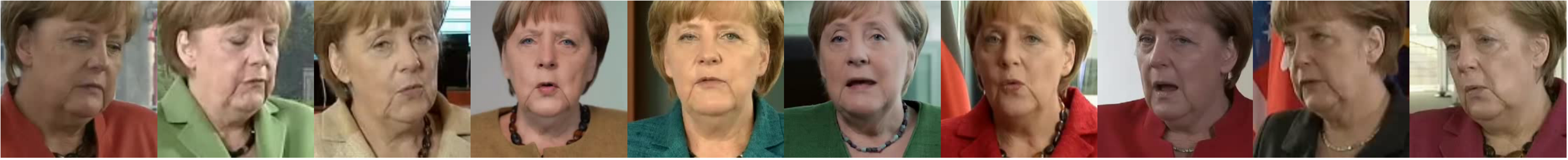}\\
    \hspace*{\fill}Figure 1: Angela Merkel looking in all directions during video podcasts.\hspace*{\fill}
 }
\begin{document}

\maketitleabstract%



\section{Introduction}

Multi-modal spoken video corpora are becoming more widespread and relevant in recent years. 
Given today's compute power, multi-modal analyses as well as cross-modal learning between video and audio -- and particularly: between facial mimicry and speech -- have become more commonplace. 
For example, based on weekly addresses of then-president of the United States, Barack Obama
, ObamaNet \cite{kumar2017obamanet} produces speech and lip-synchronized facial videos from text and some intermediate representation.

There are a number of spoken video corpora available for end-to-end audio-visual machine learning tasks. The LRS~2 \citelanguageresource{Chung_2016} and LRS~3 \citelanguageresource{afouras2018lrs3ted} datasets contain thousands of sentence-like units from many different speakers collected from the BBC and from TED talks, i.\,e.\ spoken \emph{in the wild}.

The broadness and wide coverage of such multi-spea\-ker corpora come with some limitations. In particular, speech samples are devoid of their original context as short snippets are available only in isolation. Very little material is available per-speaker and snippets tend to be short and typically contain only single phrases. This prohibits modelling long-range phenomena such as conversational prosody and the focus on short snippets makes it impossible to model speaker behaviour during short speech pauses.

On the other hand, there is also already a number of single-spea\-ker video corpora collected from speeches of public figures, 
such as former US presidents Obama \citelanguageresource{janssoone:hal-02873020} and Trump \citelanguageresource{ruf-navarretta-2020-creating}. These typically are purely single-spea\-ker and do not contain any material from other speakers. Also, prepared speeches do not necessarily contain the full prosodic variability of spontaneous speech.

An important shortcoming of the existing corpora is the lack of multi-lingual data from the same speakers. For example, to test automated approaches to lip-synchronous dubbing \cite{chaume2012audiovisual,saboo-baumann-2019-integration}, one needs training data in one language and data to be dubbed in another. 
Current corpora do not provide speech from one speaker speaking multiple languages and can typically not be extended to cover this: in the case of LRS corpora, the speakers are not known and finding more material from these speakers is hence tedious. In the case of corpora from public figures, they typically do not speak in other languages than their mother tongue (which tends to be English). 

\begin{figure*}[t]
    \centering
    \includegraphics[width=\textwidth]{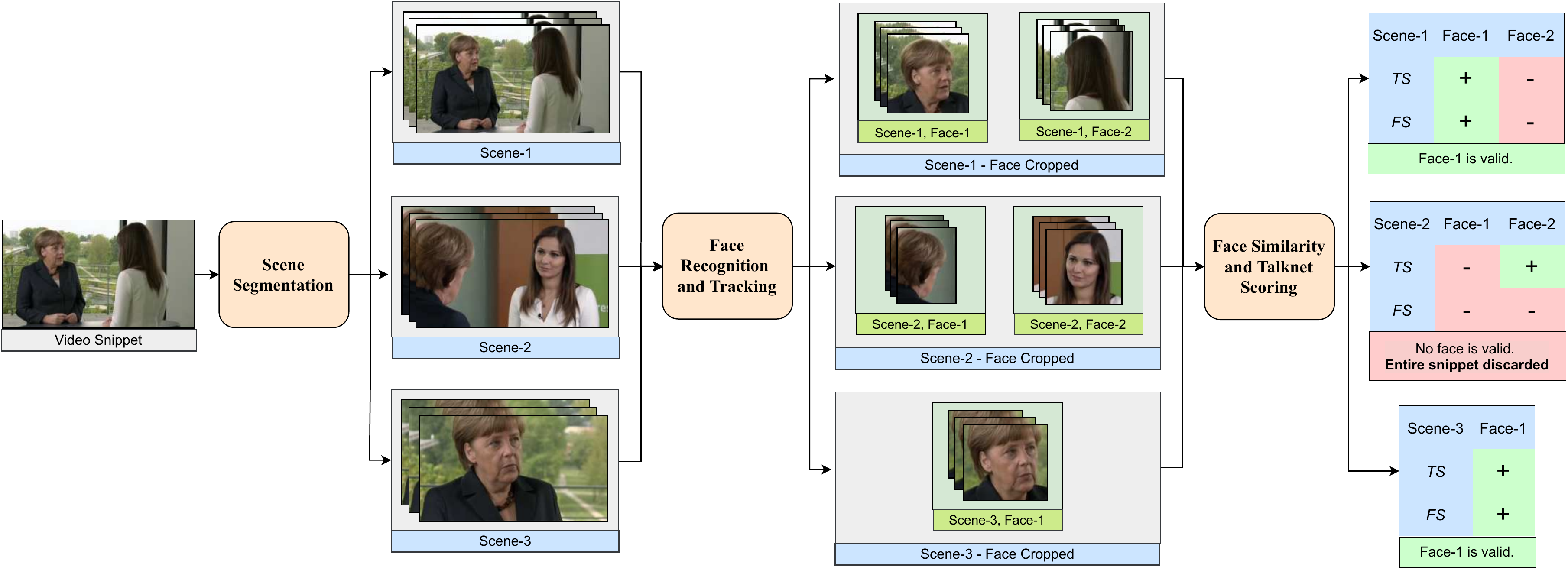}
    \caption{%
    Example of semi-automatic labeling for a target speaker (who in this case is not speaking but listening to the interviewer) in the diarization pipeline. We first segment the video into scenes. For all scenes, the faces present are recognised, tracked and cropped into smaller videos. We rate the faces in each scene using TalkNet (speaking?) and face similarity (target?) scores. We consider the best candidate face for each scene and combine them to the full face-cropped video. If no face is found valid for all scenes, the snippet is discarded (as in this case).}
    \label{fig:dubbingdatapipeline}
\end{figure*}

We introduce the Merkel Podcast Corpus.  
This corpus (a) contains large amounts of speech from one public figure (former German chancellor Angela Merkel) to the extent that it can meaningfully be used to train modern single-speaker deep learning models; 
(b) also contains further material from many other speakers (interviewers) which can help generalize to multi-spea\-ker models;
(c) contains material that is both closely aligned (short snippets of video with the corresponding texts) and consecutive in nature, i.\,e., snippets can be contextualized into the overall situation. The data spans 16 years and the primary speaker's voice ages correspondingly while recording technology advances; 
(d) comes with large amounts of meta-data available, such as date of recording, speaker names and further written documentation that the speech material can be related to;
(e) is amended with the few publicly available video recordings of Angela Merkel speaking English, which is particularly useful for cross-lingual processing tasks such as evaluating lip-synchronous dubbing.

Our second contribution is a method for semi-auto\-ma\-ti\-cally differentiating speakers and tuning in on one target speaker in collections of multi-speaker videos based on a few examples of the target speaker.
The raw material that our corpus is built from is mostly interview-style video in which the view often shifts between interviewer and interviewee. 
Often these shifts do not match speaker change. 
Our processing pipeline uses face detection and recognition, as well as speaking-face detection in order to find video snippets in which the target speaker is both visible and speaking. This processing pipeline can be used to create single-speaker corpora from multi-speaker sources (such as singling out the host or recurring guests from many episodes of a talk show). 

We describe our corpus extraction and preparation pipeline in Section~\ref{sec:pipeline} and then describe the resulting dataset in Section~\ref{sec:dataset}. 
We believe the corpus can be useful in a variety of ways and we describe some current 
usages in Section~\ref{sec:applications} and conclude in Section~\ref{sec:conc}.

\section{Dataset Source and Preparation}
\label{sec:pipeline}

Angela Merkel decided early on in her tenure to produce weekly video podcasts to be published via Internet. More than 600 such podcasts have been published between 2006 and 2021, yielding much more material and covering a longer time than Obama or Trump. 
Although always high-quality productions, video quality and resolution has substantially evolved over time. Since 2013, podcasts contain lightly edited subtitles; older videos are amended at least with a written transcript for accessibility. 
The format changed from semi-spontaneous speeches to 
predominantly interviews in 2011-2018. 

\subsection{Scraping, Alignment and Snippeting}

We scraped the official video platform to download all videos including subtitles, transcripts and other alternative formats as available.\footnote{See \scriptsize\url{https://www.bundesregierung.de/breg-de/service/archiv/archiv-podacasts}.} 
After extracting the text from transcript PDFs, we use robust forced alignment \cite{swcLRE2018} to align text and speech. 

Cross-modal learning applications typically require many short audio-video-text snippets rather than few long files. 
We built flexible snippeting scripts that consider linguistic units and use speech pauses as possible cuts and merge together adjacent material up to a given threshold (to match GPU memory constraints). It is possible to overlap snippets for data augmentation and the minimal units (words, inter-pausal-units, sentences) can be set flexibly. 
Alternatively, snippets can be created from subtitle files. (See Section~\ref{sec:dataset} for a comparison of alignments and subtitles.)

\subsection{Speaker Diarization}

Audio-visual corpora should contain video where the face of the target person speaking is visible and she is actually speaking. In the raw videos, however, a different person may be speaking (with the target speaker visible or not), or the target speaker may be speaking but not visible. This is not limited to interviews but can also be caused by pre-recorded video footage interspersed in single-speaker podcasts. Furthermore, for multiple faces visible on screen, the target speaker's face should be identified.

Our pipeline for identifying snippets of the target person is depicted in Figure~\ref{fig:dubbingdatapipeline} and works as follows:
For each snippet, we identify and temporally crop scene/camera changes using PySceneDetect\footnote{\scriptsize\url{https://github.com/Breakthrough/PySceneDetect}} and then use S\textsuperscript{3}FD \cite{zhang2017s3fd} to detect and crop faces in the extracted scenes.\footnote{These steps are inspired by SyncNet \cite{Chung16a} preprocessing.} To get the lips we crop the bottom half of the detected faces. 

Snippets must show the target person and that person must be speaking.
We perform two checks on the face-cropped scenes:
\begin{compactitem}
\item We use the \emph{Talknet} active speaker detection tool\footnote{\scriptsize\url{https://github.com/TaoRuijie/TalkNet\_ASD}} \cite{2021} to detect whether the speaker on-screen is talking. Regardless of who the target speaker(s) are, this is necessary to ensure that the lip movements visible in the face-cropped video matches the speech segment. 
\item We check whether the person on screen is one of the target speakers using a face recognition tool\footnote{\scriptsize\url{https://github.com/ageitgey/face\_recognition}} that checks similarity with a reference set of images from known target-speaker videos.\footnote{We manually selected 20 snippets of Merkel as samples for face recognition.}
\end{compactitem}
A positive result in both of these tests ensures that our target speaker is speaking on-screen in the face-cropped scene. 
%
If, for some scene, none of the cropped faces pass the above two checks, the entire snippet is discarded. Otherwise, the best face-cropped and lip-cropped candidates for each scene are recombined to give the face-cropped and lip-cropped videos respectively. The pipeline is demonstrated with an example in Figure \ref{fig:dubbingdatapipeline}. On a manually annotated validation set containing 100 ground truth examples, the pipeline achieves an accuracy of 94\,\% with perfect precision, ensuring no false positives in the processed output.

\section{Dataset Statistics}
\label{sec:dataset}

\begin{figure}[t]
    \centering
    \vspace*{-1.8em}
    \includegraphics[width=\linewidth]{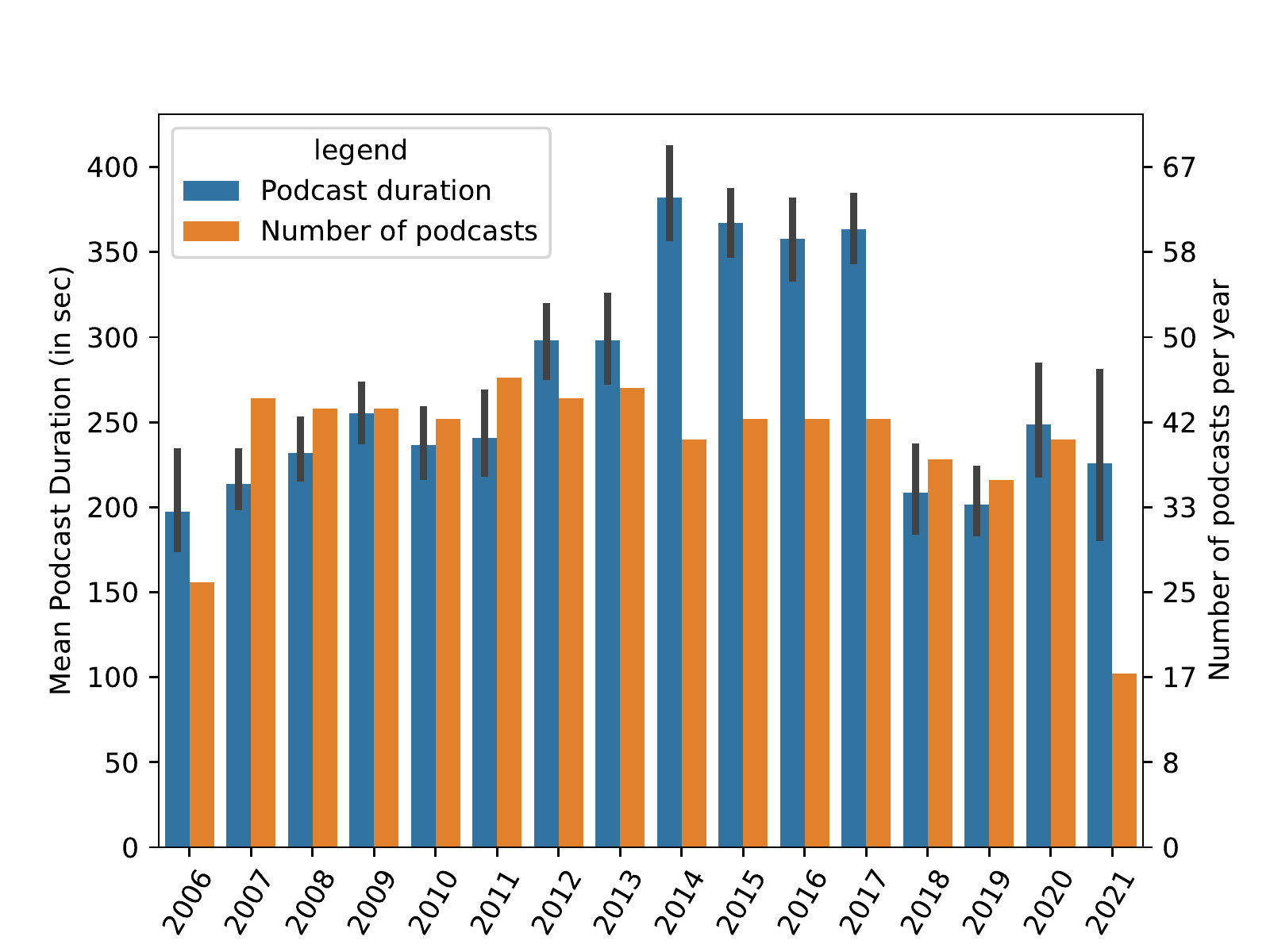}
    \vspace*{-2.2em}
    \caption{Number of podcasts and mean durations (with stddev as error bars) over time.}
    \label{fig:podcaststatsoverview}
\end{figure}

The overall corpus consists of 630 videos 
totalling 48.0\,h of material and distributed over time as shown in Figure~\ref{fig:podcaststatsoverview}. Leading and trailing jingles make up 2.7\,h and there are 2.8\,h of (short) silence, leaving 42.5\,h of speech according to the INA speech segmenter\footnote{\scriptsize\url{https://github.com/ina-foss/inaSpeechSegmenter}} \cite{ddoukhanicassp2018}.

Among the podcasts, there are more than 250 interviews, almost all with different 
interviewers that at least ask some questions, providing for a rich multi-speaker background beyond Angela Merkel as the target speaker.
The frequency of podcasts has remained stable over the years; the average duration increased to about 5-6 minutes during the `interview years' 2012-17, from 3-4 minutes before and after.

All audio is encoded as high-quality AC3 and video resolution was increased to HD in 2018 (at the same time the audio bitrate reduced from 350kbit/s, presumably because of adaptive encoding). The signal-to-noise ratio is excellent for speeches and still good for interviews. 

Angela Merkel appears on-screen for approximately 66\,\% of the time, and was found to be the on-screen active speaker for around 58\,\% (statistics obtained from face recognition and active speaker detection which both are tuned for precision). 
As a corollary, the corpus contains 3.8\,h of video of Merkel listening to someone. 

Forced alignment (FA) generally is inferior to subtitle timings, although both have their advantages. 
Our FA errs on the side of quality (over quantity) and aligns only about 62\,\% of all words; frequently, the beginning or end of sentences is missing (only 54\,\% of sentences have the first and last words aligned). Furthermore, we discard sentences where too many words are missing in the middle. We find that the mean absolute error of subtitle timings is 280\,ms and large differences ($>$1\,s) are rare. However, we also notice that subtitles (and also transcripts) are often mildly edited (changes in word order, repetitions, \ldots) or even completely wrong. FA helps to find and fix these. Our manually corrected transcripts will be made available with the corpus.

\begin{figure}[t]
    \centering
    \vspace*{-2.5em}
    \includegraphics[width=\linewidth]{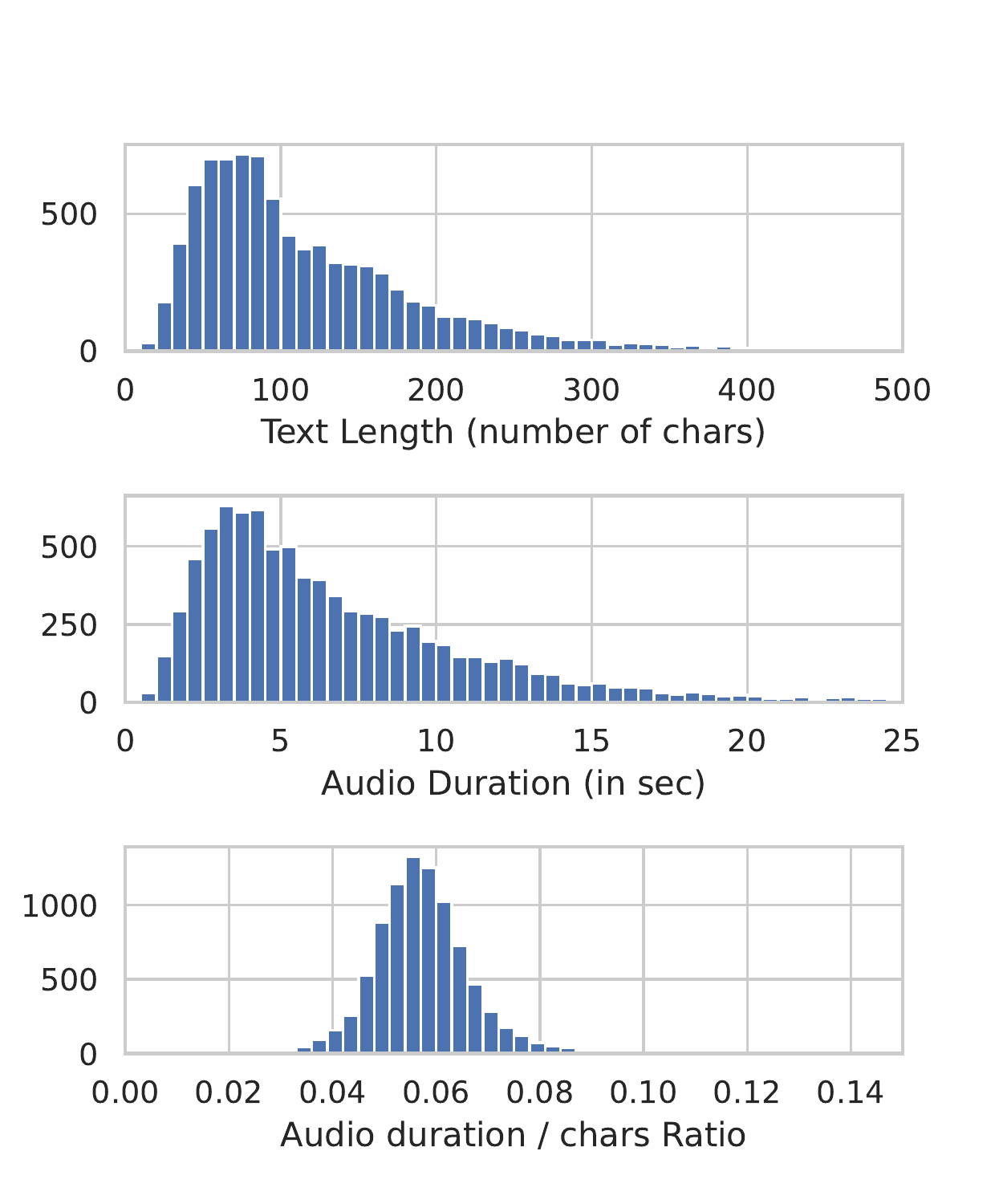}
    \vspace*{-3em}
    \caption{Target Speaker Dataset Statistics:
We show per-snippet histograms of text length, audio duration and tempo as expressed by the ratio of duration to text.}
    \label{fig:singlespeakerstats}
\end{figure}

\begin{figure}[b]
    \centering
    \includegraphics[width=\linewidth]{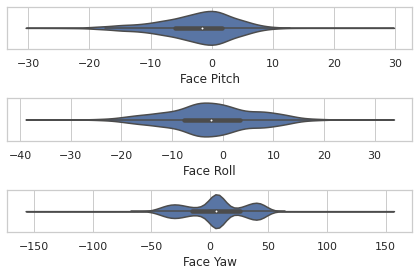}
    \vspace*{-2em}
    \caption{Violin Plots of the face angles (in degrees) present in the target-speaker corpus. Two additional peaks in the face yaw-plot belong to the left and right facing videos.}
    \label{fig:face}
\end{figure}

Based on the analysis, we decide to construct snippets for the single-speaker corpus based on subtitle files and to back off to alignments for podcasts before 2013 (where subtitles are not available).
The resulting snippets (Angela Merkel visibly speaking in video, audio and with corresponding text) make up the single-speaker corpus that can be used for ML applications (see next section) and is described further in the next paragraphs. Note, though, that the corpus can also be snippeted in different ways and contextualization is always possible through the correspondence to the source videos. 

Snippets have a mean duration of 7.2\,s (std\,dev 6.2\,s) and mean text length of 124 characters (std\,dev 104 characters). Distributions are shown in  Figure~\ref{fig:singlespeakerstats}. The third subfigure furthermore shows that speech tempo (and pausing) varies between snippets. 


Figure~\ref{fig:face} shows violin plots of facial pose. Overall, the variability in the corpus is similar to LRS\,3, although on average, Merkel raises her chin slightly more, rolls her face more and the interview-style camera positions result in distinct yaw peaks. Yaw angles are also shown in Figure~1.

\subsection{Supplementary English Speech}

In addition to the podcasts, we acquired three videos of Merkel speaking English. All three are introductions to speeches with the main content in German (before the US Congress, the British Parliament and at Stanford). 
For some of these, 
official translations are available.

We manually aligned her English utterances which amount to 86 snippets containing 250\,s of speech (excluding pauses). While too little for machine learning, it might suffice for cross-lingual evaluation. 

\section{Machine-learning Applications} \label{sec:applications}

In this section, we show different preliminary usages of the single-speaker corpus that spans a long time period and contains the visual modality. 
Our examples are by no means meant to cover the corpus' usage completely. 

\subsection{Age estimation}

Given speech from one speaker over 16 years, we wondered if timing/aging effects can be observed. We extract pre-trained speaker embeddings\footnote{\scriptsize\url{https://github.com/RF5/simple-speaker-embedding}} 
from each snippet and train a simple softmax logistic regression network to estimate what year the recording was made. For this 16-class problem, we get a macro-f1 of .63 (and regression coefficient of $R^2=.77$). This indicates that the learnt embeddings -- although they are supposed to be speaker-specific -- change with age of the speaker and/or recording. 
We intend to further research automatic age estimation which so far has exclusively been performed on multi-speaker databases.

\subsection{Lip generation} \label{w2l}

In lip generation, the task is to re-create lip movements in video when revoicing it with different speech (e.\,g.\ coming from TTS). 
We re-trained Wav2Lip \cite{2020} on the single-speaker corpus and we find that lip renderings (and particularly the movements that result in video) look much more natural. In particular, Merkel tends to not open her mouth as far as off-the-shelf Wav2Lip models would and this is fixed by a speaker-specific model.

\subsection{Visually Grounded Speech Synthesis}
The task of synthesizing speech based on text and video (to make the speech appear in sync with the facial movements) has recently gained traction \cite{hassid2021words,lu2021visualtts,hu2021neural}. 
This is particularly important for revoicing video in another language (dubbing) where it avoids phantom effects (lips moving without speech or vice-versa). It can thus be seen as an alternative to lip generation to yield synchronized video.

We train a visually grounded text-to-speech system based on Tacotron-2 \cite{shen2018natural} 
which we augment by adding an attention mechanism over video.
In preliminary experiments we find that synchrony often improves drastically. However, the challenges of the Merkel Corpus, in particular the much longer snippets, the prosodic structures and pauses contained, sometimes lead to 
erratic behaviour. 

\section{Conclusion}\label{sec:conc}

We have presented a multi-modal corpus of semi-prepared speeches and more spontaneous interview-style speech with one primary speaker (speaking for 28\,h excluding silences) and more than 250 secondary speakers (speaking for another 14\,h) over 16 years of production of a weekly podcast series.
We supplement this corpus with some English speech from the primary speaker in the hope that it may be useful for research in lip-synchronous audio-visual translation.

We have presented some preliminary results that show that the corpus can be used for state-of-the-art ML tasks in which it shows promising results. 
Beyond machine-learning applications, the corpus could be useful in the Digital Social Studies and Political Sciences as it makes available sixteen years of German federal policy and politics in a way that can easily be automatically analyzed and searched in a uniform way. 

The German government -- unlike e.\,g.\ the US government -- does not place its works in the public domain. However, recent changes in German copyright law \cite{meyer2018new} allow scraping publicly available data for research purposes and we may share the data with other researchers. 
As the videos and transcripts remain available from the original source, we provide only the speech-text alignments as well as the snippeting that we used in our experiments.  
We have also presented a simple pipeline for multi-modal speaker diarization in multi-face video which simplifies the conversion of large video collections (e.\,g.\ from talk shows) to single-speaker corpora suitable for cross-modal machine-learning tasks. We publicly release the scripts required to download and process the corpus as well as the annotations that we derived from the original data.\footnote{Download scripts and annotations are made available at \scriptsize\url{https://github.com/deeplsd/Merkel-Podcast-Corpus}.
The original video data is available upon e-mail request from the third author.}

Finally, ethical implications must be considered when dealing with so much data from one person -- or even trained models for personalized visually grounded speech synthesis. Given that Angela Merkel is a public figure who is no longer in office, we believe that the public interest outweighs the possible impact on her.

\section*{Acknowledgements}
We wish to thank Juliane Röscheisen for helping with the scraping scripts 
used to download the corpus and 
Christian Schuler for valuable feedback.
The authors would like to thank the German Academic Exchange Service  DAAD for the internship opportunity that made this work possible.

\section{Bibliographical References}\label{reference}

\bibliographystyle{lrec2022-bib}
\bibliography{lrec2022-example}

\section{Language Resource References}
\label{lr:ref}
\bibliographystylelanguageresource{lrec2022-bib}
\bibliographylanguageresource{languageresource}

\end{document}